\documentclass{article}

    \PassOptionsToPackage{numbers, compress}{natbib}

\usepackage[preprint]{neurips_2022}

\usepackage[utf8]{inputenc} 
\usepackage[T1]{fontenc}    
\usepackage{hyperref}       
\usepackage{url}            
\usepackage{booktabs}       
\usepackage{amsfonts}       
\usepackage{nicefrac}       
\usepackage{microtype}      
\usepackage{xcolor}         
\usepackage{subcaption}
\usepackage{natbib}
\usepackage{graphicx}
\usepackage{svg}
\usepackage{hyperref}
\hypersetup{
    colorlinks=true,
    citecolor=black,
    linkcolor=red,
    urlcolor=cyan
}

\title{Grokking as Compression: A Nonlinear Complexity Perspective}

\author{%
  Ziming Liu\thanks{Equal contribution} \\ 
  MIT \& IAIFI \\
  \texttt{zmliu@mit.edu} \\
  \And
  Ziqian Zhong$^*$ \\
  MIT \\
  \texttt{ziqianz@mit.edu} \\
  \AND
  Max Tegmark \\
  MIT \& IAIFI \\
  \texttt{tegmark@mit.edu} \\
}

\usepackage{graphicx}
\usepackage{dcolumn}
\usepackage{bm}
\usepackage{float}
\usepackage{subcaption}
\usepackage{multirow}
\usepackage{comment}
\usepackage{dsfont}
\usepackage{amsthm}
\usepackage{xcolor}
\usepackage{array}
\usepackage{eucal}
\usepackage{makecell}
\usepackage{mathtools}
\usepackage{makecell}

\usepackage{enumitem}
\setlist{leftmargin=10mm}

\newcommand{\mat}[1]{\mathbf{#1}}

\def\spose#1{\hbox to 0pt{#1\hss}}
\def\simlt{\mathrel{\spose{\lower 3pt\hbox{$\mathchar"218$}}
     \raise 2.0pt\hbox{$\mathchar"13C$}}}
\def\simgt{\mathrel{\spose{\lower 3pt\hbox{$\mathchar"218$}}
     \raise 2.0pt\hbox{$\mathchar"13E$}}}
\def\simpropto{\mathrel{\spose{\lower 3pt\hbox{$\mathchar"218$}}
     \raise 2.0pt\hbox{$\propto$}}}

\def\beq#1{\begin{equation}\label{#1}}
\def\eeq{\end{equation}}
\def\beqa#1{\begin{eqnarray}\label{#1}}
\def\eeqa{\end{eqnarray}}

\begin{document}

\maketitle

\begin{abstract}
     We attribute grokking, the phenomenon where generalization is much delayed after memorization, to compression. 
     We define \textit{linear mapping number} (LMN) to measure network complexity, which is a generalized version of linear region number for ReLU networks. LMN can nicely characterize neural network compression before generalization. Although $L_2$ norm has been popular to characterize model complexity, we argue in favor of LMN for a number of reasons: (1) LMN can be naturally interpreted as information/computation, while $L_2$ cannot. (2)  In the compression phase, LMN has nice linear relations with test losses, while $L_2$ is correlated with test losses in a complicated nonlinear way. (3) LMN also reveals an intriguing phenomenon of the XOR network switching between two generalization solutions, while $L_2$ does not. Besides explaning grokking, we argue that LMN is  a promising candidate as the neural network version of the Kolmogorov complexity, since it explicitly considers local or conditioned linear computations aligned with the nature of modern artificial neural networks. 
\end{abstract}




\section{Introduction}

Grokking, the phenomenon where generalization happens long after memorization~\cite{power2022grokking}, is challenging our understanding of deep learning. Although there have been a few seemingly independent explanations of grokking~\cite{liu2022towards, nanda2023progress, liu2022omnigrok,merrill2023tale,barak2022hidden,davies2023unifying,thilak2022slingshot,gromov2023grokking,notsawo2023predicting,varma2023explaining}, many of them share a similar high-level idea which is "grokking is compression": There exist a generalization solution and a memorization solution; the memorization solution is easier to be learned so learned at first, but the  generalization solution is more efficient so emerges later. Although various measures have been proposed to characterize the process of "compression", e.g., $L_2$~\cite{liu2022omnigrok}, Fourier gap~\cite{barak2022hidden}, network efficiency~\cite{varma2023explaining}, neither of these measures admits a natural interpretation as information/computation complexity (most are, at best, proxies).

\begin{figure}[tbp]
    \centering
    \includegraphics[width=0.95\linewidth]{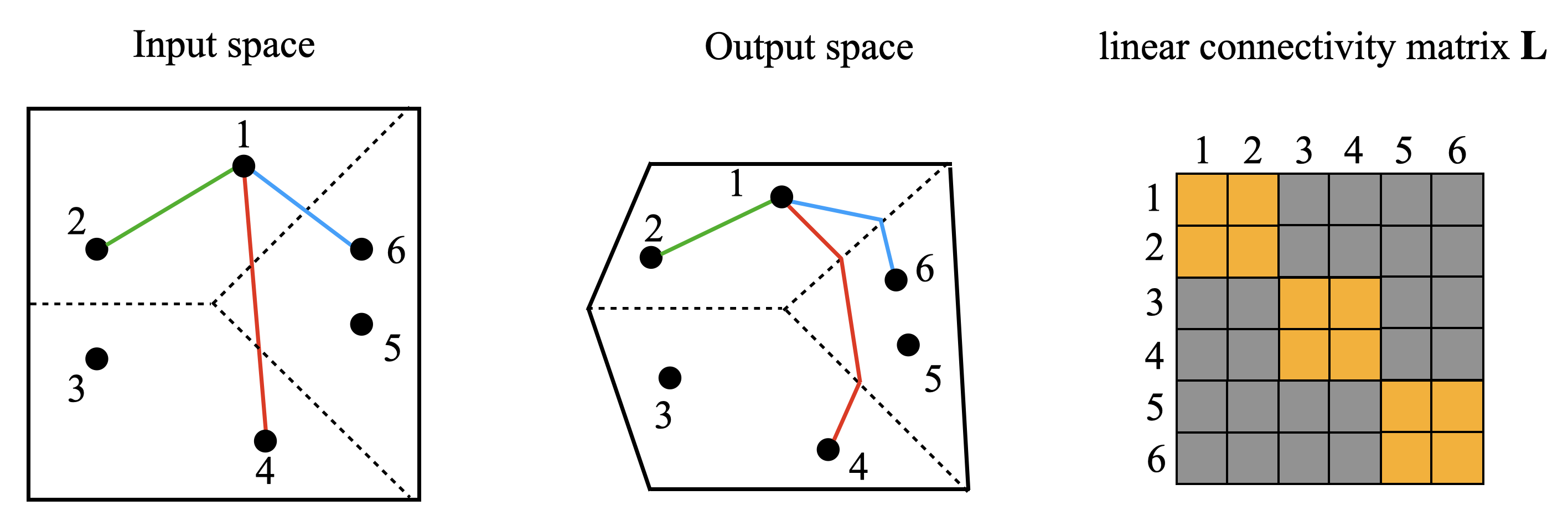}
    \caption{Linear mapping number (LMN) is a generalized version of linear region number for ReLU networks. A ReLU network partitions input space into piece-wise linear regions. If two points lie in the same linear region or different linear regions, the line connecting them in the input space (left) will remain linear (green) or turn into non-linear curves (red and blue) in the output space (middle). We can construct a linear connectivity matrix (right) to characterize whether two points lie on the same linear region, which is applicable to networks with any activations. Based on the Von Neumann entropy of the matrix, we can estimate the number of linear mappings (details in Section~\ref{sec:lmn}). }
    \label{fig:illustration}
    \vspace{-2mm}
\end{figure}

We propose a metric called \textit{linear mapping number} (LMN), which measures the complexity of a network (or a subnetwork). In brief, LMN is a generalized version of the linear region number for ReLU networks. ReLU networks are known to represent piecewise linear functions;  they partition input space into regions on which the network is a local linear mapping; different regions have different linear mappings, as shown Figure~\ref{fig:illustration}. Geometrically, one can think of ReLU networks as origami, i.e.,  folding flat input space (Figure~\ref{fig:illustration} left) into complicated shapes (Figure~\ref{fig:illustration} middle), and the number of linear regions measures the network complexity. LMN generalizes the concept of linear region number to networks with smooth activations. 

We argue that LMN is a better metric than $L_2$, which has been used to measure network complexity in deep learning, especially for grokking~\cite{liu2022omnigrok}. A conceptual example is linear networks, which can only represent linear mappings even when they are deep. For linear networks, LMN always gives 1, but $L_2$ can be arbitrary hence not very informative. Moreover, LMN can be naturally interpreted as information: if one wants to compress a network into (input-dependent) linear mappings, then the compressed information is basically LMN times the size of one linear mapping.

We use LMN to characterize the compression process of grokking on three algorithmic tasks: modular additon, permutation group $S_4$ and multi-digit XOR. After memorization and before generalization, the LMN decreases steadily, and has a strong linear relation with test loss. By contrast, $L_2$ is correlated with test losses in a complicated nonlinear way. For modular addition and permutation, the LMN starts to level off after grokking, as expected. For multi-digit XOR, LMN displays an unexpected double-descent after grokking. 
This reveals something intriguing about the XOR case, which has two (rather than one) generalization solutions which are almost degenerate, so the network jumps between these two solutions. 

This paper is organized as follows: In Section~\ref{sec:lmn}, we define linear mapping number (LMN). In Section~\ref{sec:grokking}, we use LMN to explain grokking, showing that it is related to $L_2$ but also better than $L_2$ in serveral senses. We discuss related works in Section~\ref{sec:related_works}.

\section{Linear Mapping Number (LMN)}\label{sec:lmn}

The linear mapping number (LMN) is a generalization of the linear region number for ReLU networks. For simplicity, let us first consider ReLU networks. A ReLU network partitions input space into linear regions, where in each region the ReLU network behaves like a linear mapping locally, although different linear regions correspond to different linear mappings (see Figure~\ref{fig:illustration}). The number of linear regions has been proposed to measure network complexity for ReLU networks~\cite{montufar2014number,hanin2019complexity}.

While the linear region number is only defined for networks with ReLU activations, our proposed linear mapping number is defined for networks with {\it any} activations, including smooth ones. However, ReLU networks point to a route for how to define LMN generally. As illustrated in Figure~\ref{fig:illustration}, if two samples lie in the same or different linear regions, a straight line connecting them in input space (Figure~\ref{fig:illustration} left) will remain linear or become non-linear in output space (Figure~\ref{fig:illustration} middle). This inspires us to measure "linear connectivity" between two samples: The more linear the output line is, the larger the linear connectivity is. For a network $\mat{f}:\mathbb{R}^{d_1}\to\mathbb{R}^{d_2}$, and two input samples $\mat{x}^{(i)},\mat{x}^{(j)}\in\mathbb{R}^{d_1}$, $i,j\in[N]$, we denote the linear connectivity of them as $L_{ij}\in\mathbb{R}$. We interpolate linearly between $\mat{x}^{(i)}$ and $\mat{x}^{(j)}$ in input space:
\begin{equation}
    \mat{x}^{(i,j)}(\lambda) = \mat{x}^{(i)} + \lambda (\mat{x}^{(j)} - \mat{x}^{(i)}),\ \lambda\in [0,1],
\end{equation}
which corresponds to the output curve $\mat{y}^{(i,j)}(\lambda)=\mat{f}(\mat{x}^{(i,j)}(\lambda))\in\mathbb{R}^{d_2}$. The $k^{\rm th}$ dimension $\mat{y}_k^{(i,j)}(\lambda)$ is simply a scalar function of $\lambda$, so we can evaluate its linearity by doing linear regression and calculating $r^2$ (the square of the Pearson correlation coefficient). We define $L_{ij}$ as the average of $r^2$ over dimensions $k$, i.e.,
\begin{equation}\label{eq:L}
    L_{ij} \equiv \frac{1}{d_2}\sum_{k=1}^{d_2} r^2(\mat{y}_k^{(i,j)}(\lambda), \lambda).
\end{equation}
Note that $L_{ij}\in[0,1]$. The $r^2$ is measured using uniform points on $\lambda\in[0,1]$~\footnote{In practice, we use 21 uniformly spaced points on $\lambda\in [0,1]$, i.e., $\lambda=0.0,0.05,0.1,\cdots,0.95,1.0$. The $r^2$ between variable $x$ and $y$ is $r^2(x,y)=(\langle xy\rangle-\langle x\rangle\langle y\rangle)^2/(\langle x^2\rangle-\langle x\rangle^2)(\langle y^2\rangle-\langle y\rangle^2)$, where $\langle\cdot\rangle$ means averging over samples.}. When $\mat{y}^{(i,j)}(\lambda)$ is a straight line, $L_{ij}=1$; when $\mat{y}^{(i,j)}(\lambda)$ resembles a symmetric parabola, $L_{ij}=0$. We define self-connectivity $L_{ii}\equiv 1$. In summary, larger $L_{ij}$ means that the network behaves more like a linear mapping for sample $i$ and $j$ (i.e., two samples need only one shared linear mapping), while smaller $L_{ij}$ means the network behaves non-linearly in-between sample $i$ and $j$. We can stack $L_{ij}$ into a matrix $\mat{L}$ such that $\mat{L}_{ij}=L_{ij}$, and call $\mat{L}$ the linear connectivity matrix (Figure~\ref{fig:illustration} right).

If we say linearly connected samples belong to the same linear mapping, then the problem of counting linear mappings boils down to the problem of clustering: given the sample similarity matrix $\mat{L}$, how many clusters are there? Since the number of clusters is a discrete quantity and determining it may be non-robust or hyper-parameter dependent, we use a soft estimator leveraging the eigenvalue structure of the similarity matrix inspired by Von Neumann entropy \cite{von2013mathematische}. Define $\lambda_i~(i=1,\cdots,N)$ as the eigenvalues of $\mat{L}$. Note that $\mat{L}$ is symmetric ($\mat{L}_{ij}=\mat{L}_{ji}$) hence all eiganvalues are real. $\mat{L}$ is almost semi-positive definite, i.e., all eigenvalues large in magnitude are positive, but there might be a few small negative eigenvalues (see Appendix~\ref{app:lmn-sv}), which we take their absolute values. We define normalized eigenvalues $\tilde{\lambda}_i=|\lambda_i|/(\sum_{j=1}^N |\lambda_j|)$. Then we treat the normalized eigenvalue vector $(\tilde{\lambda}_1,\tilde{\lambda}_2,\cdots, \tilde{\lambda}_N)$ as a probability distribution. We define the nonlinear complexity of the distribution (measured in bits) as
\begin{equation}\label{eq:entropy-lmn}
    S_{\rm NL} \equiv - \sum_i \tilde{\lambda}_i{\rm log_2}\tilde{\lambda}_i
\end{equation}
and define the number of linear mappings LMN as
${\rm LMN} \equiv 2^{S_{\rm NL}}$.
Note that given a data set ${\bf x}^{(i)}$, the quantity $S_{\rm NL}$ defines a measure of the nonlinear complexity of {\it any} function, regardless of whether it is defined as a neural network or not, and that 
$S_{\rm NL}=0$ for any linear or affine function.

To get some intuition of the definition above, let us consider a case where there are $c$ clusters with each cluster having the equal size $N/c$, and samples are perfectly linearly connected to other samples within the cluster. In this case, $\mat{L}$ is a block-diagonal matrix with $c$ blocks ($c=3$ illustrated in Figure~\ref{fig:illustration} right), each block being an all-one matrix. The normalized eigenvalue vector is then $\tilde{\lambda}_i=1/c\ (1\leq i\leq c)$ and $\tilde{\lambda}_i=0\ (c<i\leq N)$, whose entropy is $S={\rm log}(c)$, resulting in ${\rm LMN}=c$, as expected. Note that LMN does not only apply to the whole network, but also to any sub-network. 
In particular, LMN between an intermediate layer and the output layer is of interest. 

\section{Using LMN to explain grokking}\label{sec:grokking}

\begin{figure}
    \centering
    \includegraphics[width=0.9\linewidth]{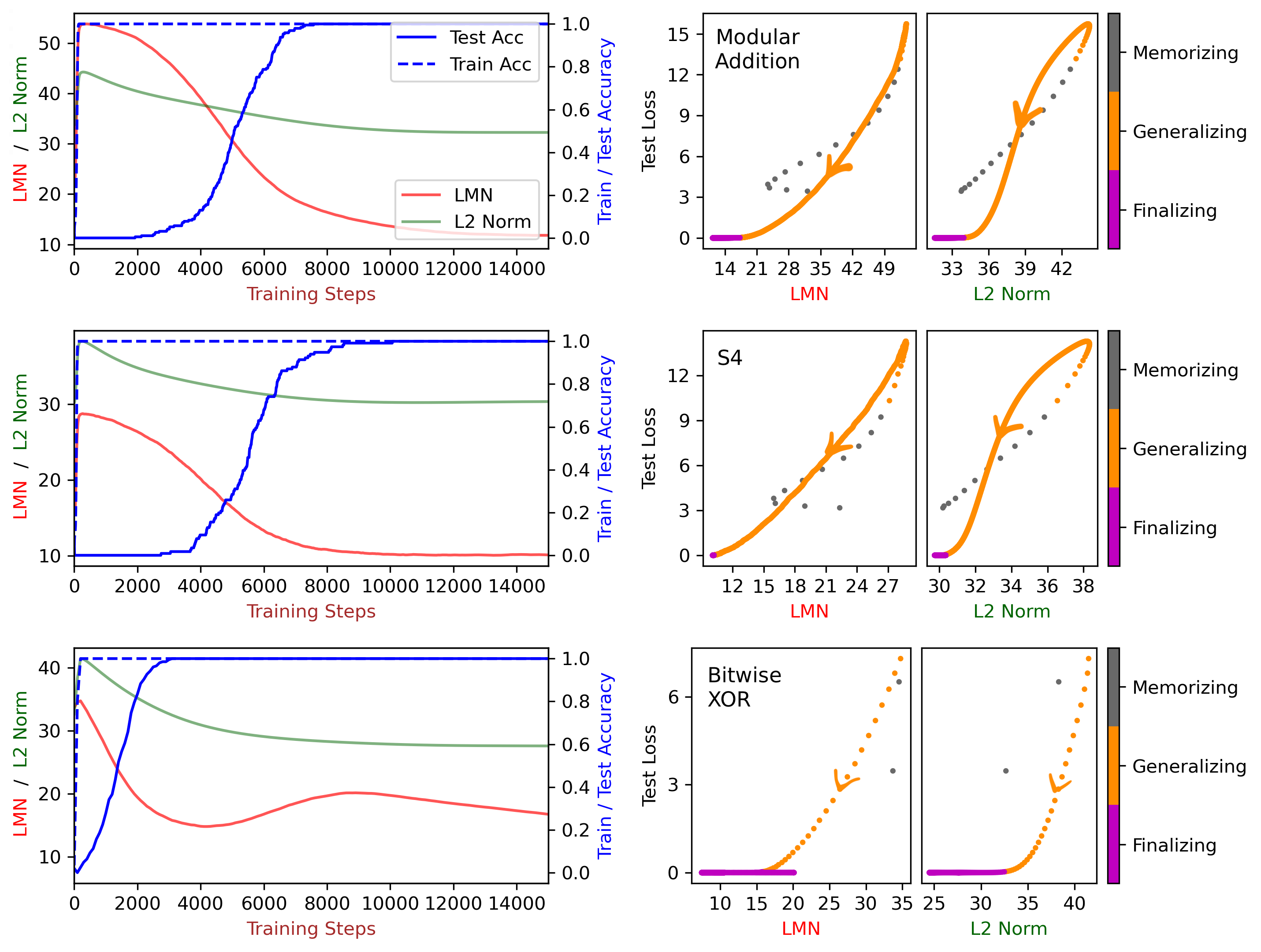}
    \caption{Train \& test accuracy, LMN (linear mapping number) after the first layer and $L_2$ norm of model parameters during the training processes. The three rows correspond to three different algorithmic tasks. \textit{Top:} Modular addition. \textit{Middle:} S4 group operation. \textit{Bottom:} Bitwise XOR.}
    \vspace{-0.5mm}
    \label{fig:results}
    \vspace{-4.5mm}
\end{figure}


In this Section, we show that LMN is able to characterize the compression process of network complexity before grokking. LMN steadily decreases between memorization and generalization. 

{\bf Experiment setup} We train three-layer fully-connected networks with SiLU activations \cite{elfwing2018sigmoid} to perform algorithmic tasks, including \{addition modulo $31$, permutation composition on $S_4$, 5-digit bitwise XOR\}. The neural network parameters (including embeddings) are trained with the AdamW optimizer (learning rate $10^{-3}$, weight decay 0.2) on cross-entropy loss for 20000 steps. The embedding dimension is 32, the hidden dimension is 100, and the output dimension is \{31, 24, 32\}. An 80-20 train-test split is performed on all possible inputs.

{\bf Results} LMN is measured between the first hidden layer and the output logit layer~\footnote{The first hidden layer is the most meaningful one for a three-layer network. The results for the embedding layer and the second hidden layer are shown in Appendix~\ref{app:lmn-all}.}. In Figure \ref{fig:results}, we plotted the LMN 
and losses during the training course for the three tasks. We denote the period before training accuracy reaches 100\% (overfitting point) the memorizing phase, the period after that but before testing accuracy reaches 100\% (generalizing point) the generalizing phase, and the remaining period finalizing phase. We see that the LMN decreases during the generalizing phase, revealing the "hidden" compression process of the network. Furthermore, the LMN is more linearly correlated than the test loss comparing to the $L_2$ norm of the model parameters.



{\bf An intriguing phenomenon in XOR} In the 5-digit bitwise XOR task, we discovered a previously undescribed phenomenon: the LMN formed a double-descent-like shape during the finalizing phase; the LMN increases briefly after generalization before decreasing again. We believe the phenomenon is due to two possible solutions for handling individual bits: we could create mapping for all the four possible pairs $(0,0), (0,1), (1,0), (1,1)$, or reduce the number of pairs to three by symmetry (handling $(0,1)$ and $(1,0)$ identically). While the latter is more efficient in terms of internal representations, the former could produce better results earlier in the finalizing phases, as the model might be unable to handle symmetries perfectly. In the period where the LMN increases after generalizing, the model could be handling asymmetries in the model: adding separate treatments for $(0,1)$ and $(1,0)$ pairs, and only favoring the more symmetric treatment after that. Evidence for the explanation is that the two turning points of the LMN are 15 and 20, which happen to be $5\times 3$ and $5\times 4$ (there are 5 digits in total; for each digit, either memorize 3 samples or 4 samples). Mechanistic investigation of this phenomenon is left for future study.

\section{Related Works and Discussions}\label{sec:related_works}
{\bf Grokking} is the phenomenon where generalization happens long after overfitting~\cite{power2022grokking}. There are some attempts to understand grokking by studying toy models~\cite{liu2022towards,gromov2023grokking}, defining measures to characterize the dynamics~\cite{nanda2023progress,liu2022omnigrok,barak2022hidden,varma2023explaining,notsawo2023predicting}, and linking to double descent~\cite{davies2023unifying} and optimization~\cite{thilak2022slingshot}. This work studies grokking from computation/information complexity.

{\bf Complexity measures for deep learning} To understand why deep learning generalizes, a number of complexity measures are proposed~\cite{jiang2019fantastic,udrescu2021symbolic,raghu2017expressive}. 
From the perspective of information (the minimal number of linear mappings required to simulate the network), linear region number is used to measure complexity of ReLU networks~\cite{montufar2014number,hanin2019complexity}, and our work extends it to linear mapping number which accommodates general networks with any activation. 

{\bf Compression and deep learning} The theory of information bottleneck~\cite{tishby2000information} suggests a compression phase followed by a fitting phase, although the compression story is sensitive to technical details~\cite{michael2018on}. Recently the success of language models is also attributed to compression~\cite{deletang2023language}. We agree that the perspectives of information and compression are very likely the key to unlock generalization puzzles of deep learning, and our proposed LMN might be a useful metric in this regard. We would like to test the usability of LMN on a broad range of tasks and architectures in the future.

\section*{Acknowledgement}
ZL and MT are supported by IAIFI through NSF grant PHY-2019786, the Foundational Questions Institute and the Rothberg Family Fund for Cognitive Science. 

\bibliography{grokking_compression.bib}

\newpage
\appendix

{\huge Appendix}

\section{LMN for all layers}\label{app:lmn-all}
\begin{figure}[h!]
    \centering
    \includegraphics[width=0.8\linewidth]{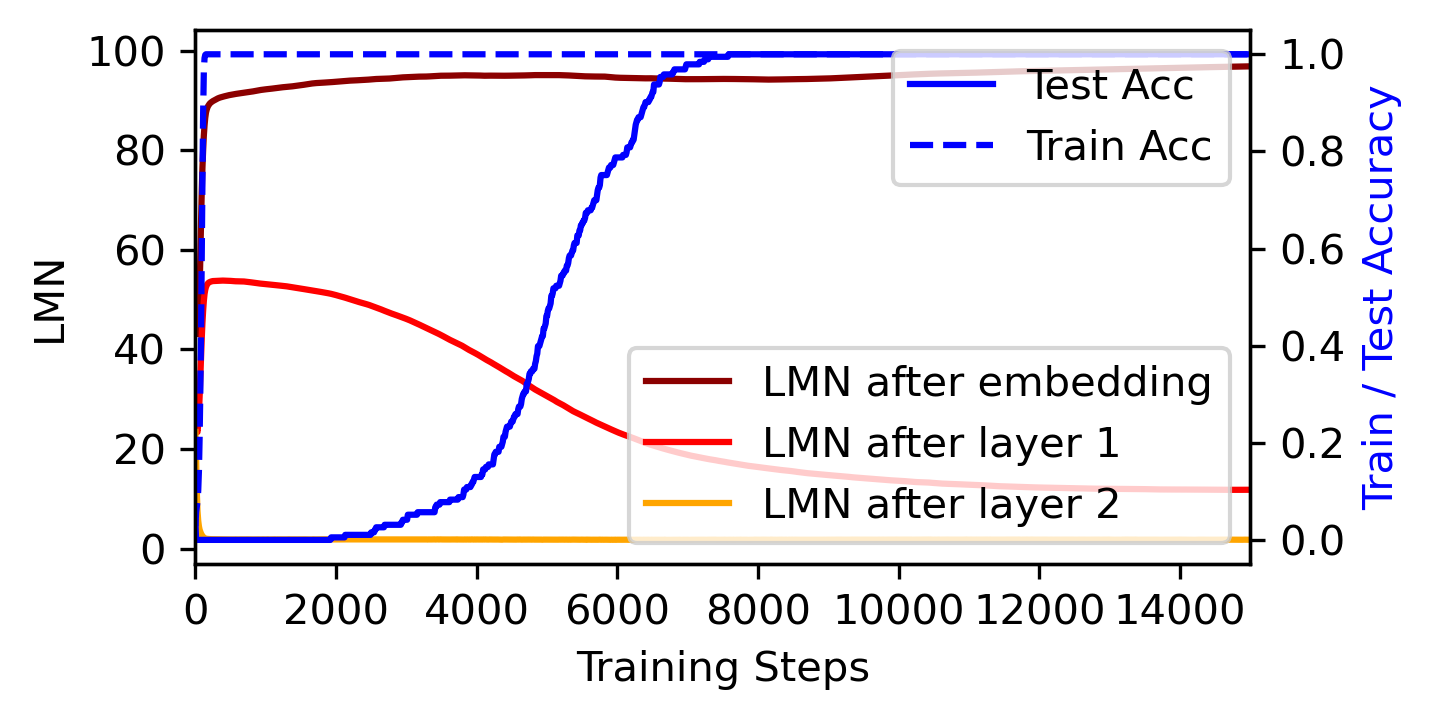}
    \caption{Evolution of LMN for all layers. Only the first hidden layer is meaningful to characterize the hidden progress before grokking, while the LMN of the embedding layer and the second hidden layer plateau quickly after memorization.}
    \label{fig:lmn-all}
\end{figure}

In Figure~\ref{fig:results}, we plotted LMN for the first hidden layer. Note that LMN can be defined for any layer, including the embedding layer and the second hidden layer. For modular addition, we show the evolution of LMN for all layers in Figure~\ref{fig:lmn-all}. It is clear that only the first hidden layer is sensitive to the hidden progress of the network after memorization and before generalization. The embedding layer and the second hidden layer are less meaningful. The embeddings are not processed by network yet, so they are not related to outputs in a meaningful way. The second layer, on the other hand, is highly correlated with the output logits, hence basically synchronizes with the training curve.

\section{Linear connectivity matrix and eigenvalue distribution}\label{app:lmn-sv}
\begin{figure}[h!]
    \centering
    \includegraphics[width=\linewidth]{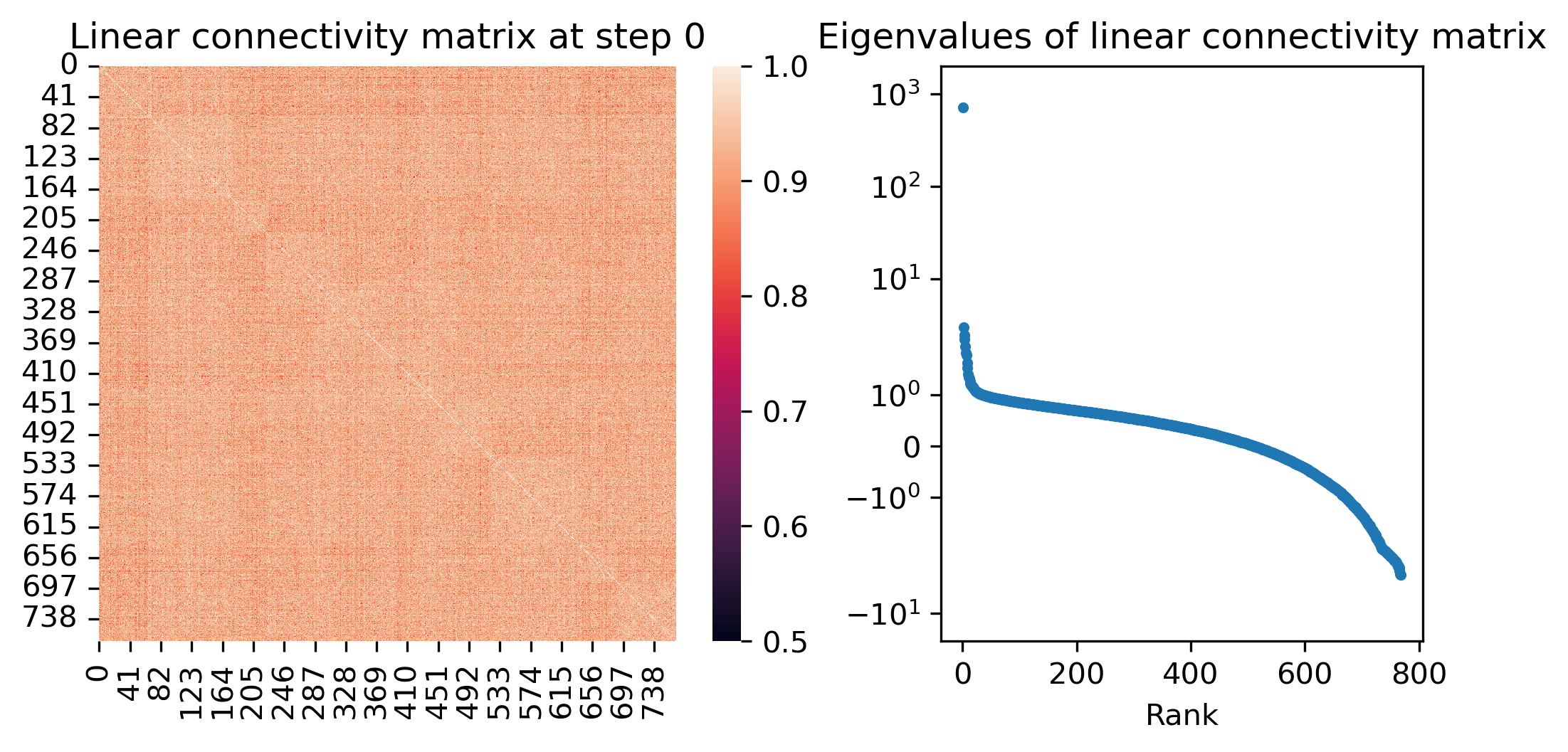}
    \includegraphics[width=\linewidth]{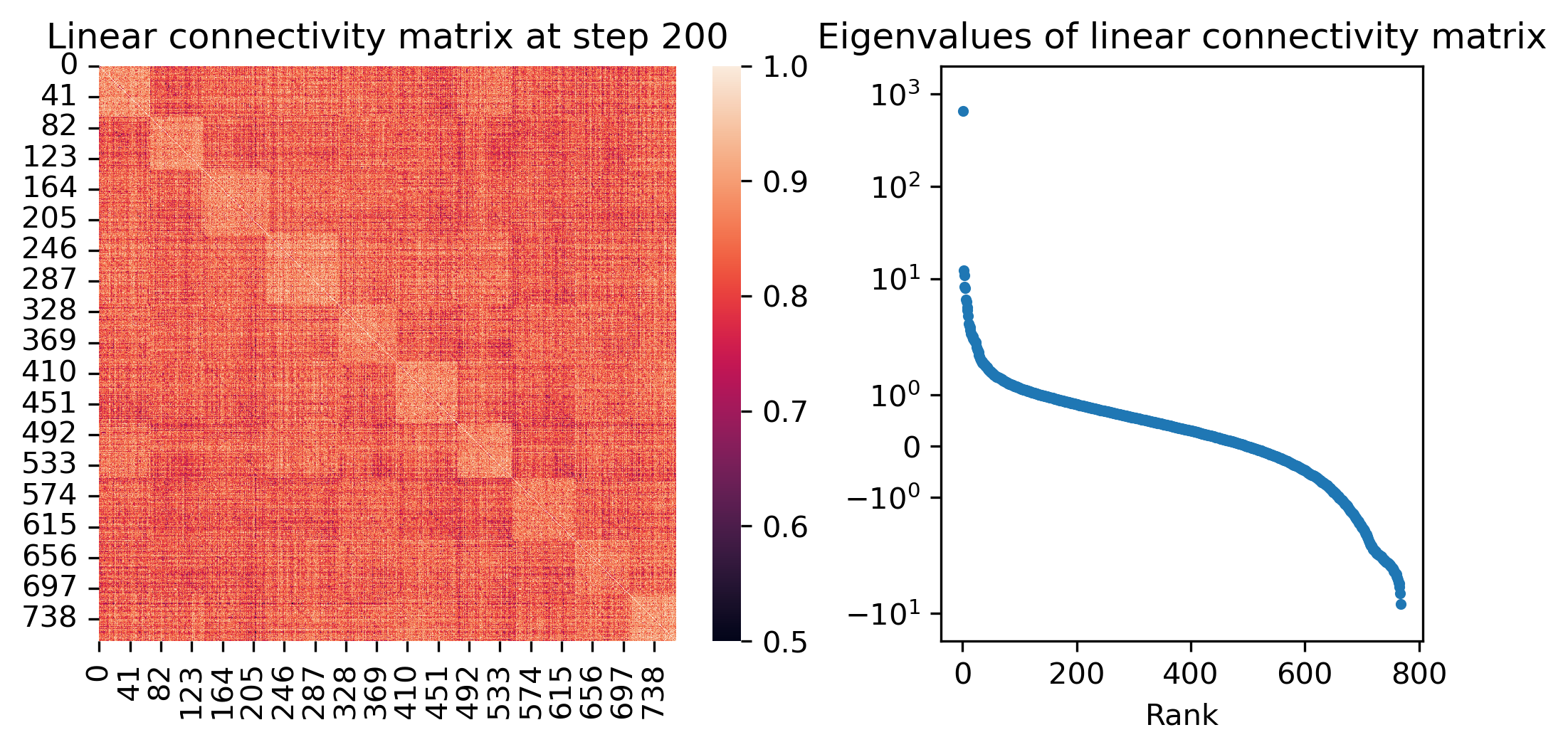}
    \includegraphics[width=\linewidth]{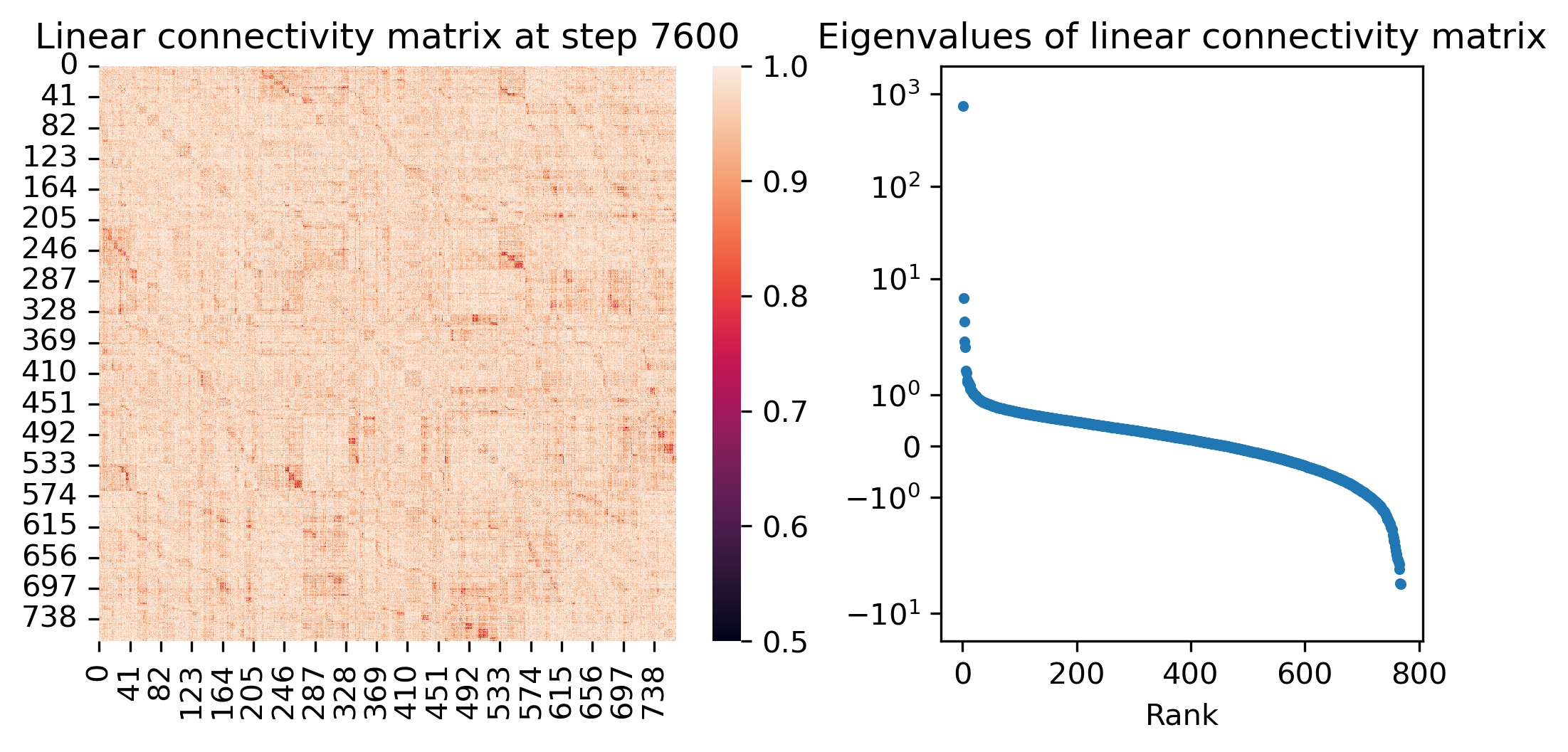}
    \caption{The evolution of the linear connectivity matrix (left) and its eigenvalues (right) at initialization (top), right after memorization (middle) and right after generalization (bottom). For display, we rearranged the input axes of the linear connectivity matrices into 10 clusters via spectral clustering (e.g. \cite{von2007tutorial}).}
    \label{fig:LCM}
\end{figure}

In the main paper, we defined linear connectivity matrix $\mat{L}$ in Eq.~(\ref{eq:L}). Here in Figure~\ref{fig:LCM}, we visualize it and show its eigenvalues for three snapshots in training (for modular addition): at initialization (step 0), memorization (step 200) and generalization (step 7600). Comparing generalization to memorization, off-diagonal elements of $\mat{L}$ are on average larger for generalization, meaning that samples are more linearly connected, hence the network is simpler for generalization. At initialization, linear connectivity is also strong, due to the simplicity inductive bias at initialization (the network is close to be a linear network at initialization).

\end{document}